\documentclass[letterpaper, 10 pt, conference]{ieeeconf}  %

\IEEEoverridecommandlockouts                              %

\usepackage{subcaption}
\usepackage{wrapfig}

\usepackage{booktabs}
\usepackage{multirow}
\usepackage[table,xcdraw]{xcolor}
\usepackage{xspace}
\usepackage{graphicx}

\usepackage{amsmath}
\usepackage{amssymb}

\usepackage{cite}

\usepackage{hyperref}

\usepackage{afterpage}

\usepackage[capitalise]{cleveref}

\newcommand{\expectation}{\mathop{\mathbb{E}}}
\newcommand{\algname}{GHIL-Glue\xspace}

\title{\LARGE \bf
\algname: Hierarchical Control with Filtered Subgoal Images
}

\author{Kyle B. Hatch$^{1}$ \and Ashwin Balakrishna$^{1}$ \and Oier Mees$^{2}$ \and Suraj Nair$^{1}$ \and Seohong Park$^{2}$ \and Blake Wulfe$^{1}$ \hspace{2em} \and
Masha Itkina$^{1}$  \and Benjamin Eysenbach$^{3}$ \and Sergey Levine$^{2}$ \and Thomas Kollar$^{1}$ \and Benjamin Burchfiel$^{1}$
\thanks{Correspondence to: \href{mailto:kyle.hatch@tri.global}{kyle.hatch@tri.global}}
\thanks{$^{1}$Toyota Research Institute}%
\thanks{$^{2}$UC Berkeley}%
\thanks{$^{3}$Princeton University}%
}

\begin{document}

\maketitle
\thispagestyle{empty}
\pagestyle{empty}

\begin{abstract}

Image and video generative models that are pre-trained on Internet-scale data can greatly increase the generalization capacity of robot learning systems. 
These models can function as high-level planners, generating intermediate subgoals for low-level goal-conditioned policies to reach. 
However, the performance of these systems can be greatly bottlenecked by the interface between generative models and low-level controllers.
For example, generative models may predict photorealistic yet physically infeasible frames that confuse low-level policies. 
Low-level policies may also be sensitive to subtle visual artifacts in generated goal images. 
This paper addresses these two facets of generalization, providing an interface to effectively ``glue together'' language-conditioned image or video prediction models with low-level goal-conditioned policies. 
Our method, Generative Hierarchical Imitation Learning-Glue (\algname), filters out subgoals that do not lead to task progress and improves the robustness of goal-conditioned policies to generated subgoals with harmful visual artifacts.
We find in extensive experiments in both simulated and real environments that \algname achieves a 25\% improvement across several hierarchical models that leverage generative subgoals, 
achieving a new state-of-the-art on the CALVIN simulation benchmark
for policies using observations from a single RGB camera.
\algname also outperforms other generalist robot policies across 3/4 language-conditioned manipulation tasks testing zero-shot generalization in physical experiments.
Code, model checkpoints and videos can be found at \href{https://ghil-glue.github.io}{https://ghil-glue.github.io}.

\end{abstract}

\section{Introduction}

As Internet-scale foundation models achieve success in computer vision and natural language processing, a central question arises for robot learning: how can Internet-scale models enable 
embodied behavior generalization?
While one approach is to collect increasingly large action-labeled robot manipulation training datasets~\cite{dasarirobonet, open_x_embodiment_rt_x_2023, khazatsky2024droid}, video datasets (without actions) from the Internet are vastly larger. 
This action-free video data can provide robotic control policies with a wide array of common sense capabilities.
However, while videos may be useful for inferring the steps in a task, such as how the objects should be moved, or which parts of an object to manipulate (e.g., grabbing a cup by the handle), they are less useful for learning details about low-level control.
For example, it is difficult to infer the action commands for controlling a robot's fingers from videos of humans performing manipulation tasks. 
One promising solution to this challenge is to employ a hierarchical approach: infer high-level subgoals in the form of goal images using models trained on Internet-scale videos, and then fill in the fine-grained motions with low-level policies trained on robot data. 

\begin{figure}[t!]
    \centering
    \includegraphics[width=0.99\linewidth]{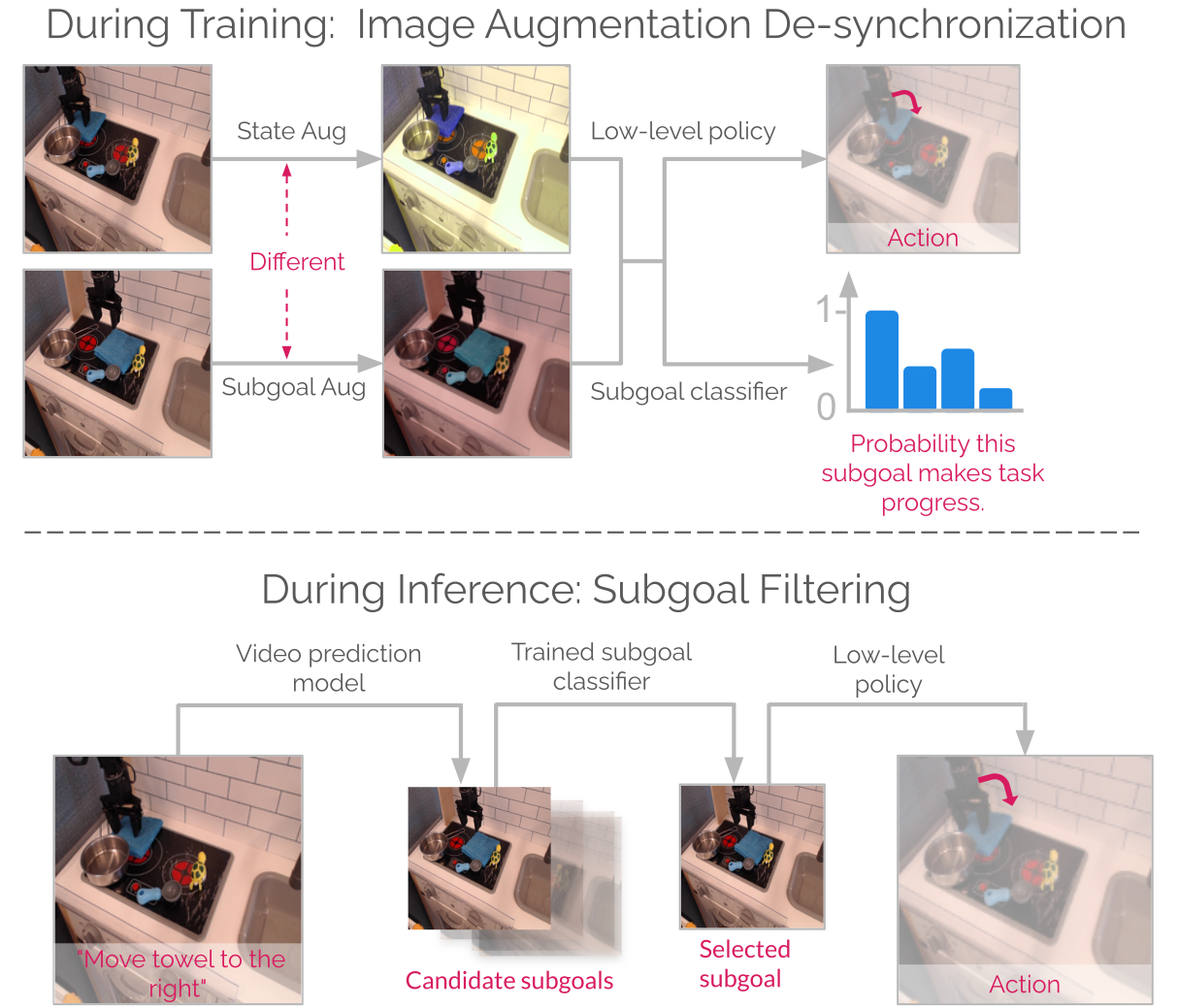}
    \vspace{-5mm}
    \captionsetup{font=small}
    \caption{
        \textbf{\algname. } We consider language-conditioned image and video prediction models that can generate multiple subgoals.
        \algname has two components: augmentation de-synchronization (top) and subgoal filtering (bottom). 
        \underline{Subgoal filtering:} We train a classifier to identify which subgoal is most likely to progress towards completing the language instruction. 
        This subgoal and the image observation are then passed to the low-level policy to choose a robot action. 
        \underline{Augmentation de-synchronization:}
        The distribution shift between subgoals sampled from the robot dataset during training and those sampled from the generative model during inference can degrade low-level policy and subgoal classifier performance.  To robustify the low-level policy and subgoal classifier to artifacts in generated subgoals, we explicitly de-synchronize the image-augmentations applied
        to the current state (State Aug) and the sampled goal (Subgoal Aug).
        }
    \label{fig:method_teaser}
    \vspace{-6mm}
\end{figure}

Modern hierarchical imitation learning algorithms~\cite{black2023zero, du2024learning} typically use an image or video generative model trained on Internet-scale data to predict subgoal images, and then use a low-level control policy to translate these subgoal images into a sequence of motor commands~\cite{black2023zero, du2024learning}. 
This approach allows the generative model 
to shoulder the hardest aspects of robotic generalization, such as generalizing to novel scenes, objects, and tasks. The low-level policy is then left with the comparatively easy task of choosing actions to reach these goals over short time horizons, which can be learned from a modest amount of robot data. 

While this general approach has seen success in prior robotic manipulation work~\cite{kapelyukh2023dall, black2023zero, du2023video, du2024learning, ajay2024compositional, gao2024can}, the interface between the high-level planner generating subgoals and the low-level policy that must reach these subgoals can be brittle.
State-of-the-art (SOTA) image or video prediction models are effective at generating likely subgoal images given a language prompt describing the task, but these subgoal generations may not be functionally useful for control.
First, generative models may occasionally sample subgoals that do not progress towards completing a given language instruction. 
If one such ``off-task'' subgoal is followed, it can have a compounding errors effect, leading to subsequent subgoals being increasingly ``off-task.''
Second, even if the generated subgoals lead to task progress, they can contain subtle visual artifacts that degrade the performance of a naively trained low-level policy.

We propose Generative Hierarchical Imitation Learning-Glue (\algname) (\cref{fig:method_teaser}), a method to \emph{robustly} ``Glue'' together image or video generative models to a low-level robotic control policy. Our method is based on two components. 
\textbf{First}, we filter out ``off-task'' subgoals that are physically inconsistent with the commanded language instruction. 
We do this by training a subgoal classifier to predict the likelihood of the transition between the current state and a given subgoal resulting in progress towards completing the provided language instruction. 
We then sample a number of candidate subgoals from the generative model and choose the subgoal with the highest classifier ranking.
\textbf{Second}, we identify a simple yet non-obvious data augmentation practice to robustify both the low-level policy and our subgoal classifier to visual artifacts in the generated subgoals.
While image augmentations are ubiquitous in robot learning methods, our key finding is that the standard way of applying image augmentations 
does not make low-level policies robust to visual artifacts in generated subgoal images.

Experiments on the CALVIN~\cite{mees_calvin} simulation benchmark and four language-conditioned tasks on the Bridge V2 physical robot platform~\cite{walke_bridgev2} suggest that \algname improves upon prior SOTA methods for zero-shot generalization while adding minimal additional algorithmic complexity. 
In both simulated and real environments, \algname achieves a 25\% improvement across several hierarchical models that leverage generative subgoals, achieving a new SOTA on the CALVIN simulation benchmark for policies using observations from a single RGB camera.

\section{Related work}

\textbf{Generative Models for Robotic Control: }
Prior works have explored diverse ways to leverage generative models,
such as diffusion models~\cite{sohl2015deep, ho2020denoising} and Transformers~\cite{vaswani2017attention}, for robotic control.
They have employed highly expressive generative models, potentially pre-trained on Internet-scale data,
for low-level control~\cite{brohan2022rt, chi2023diffusion, brohan2023rt,octo_2023,Doshi24-crossformer,Zawalski24-ecot,dasari2024ditpi},
data augmentation~\cite{mandi2022cacti, chen2023genaug, yu2023scaling},
object detection~\cite{stone2023open, peng2024learning},
semantic planning~\cite{huang2022language, huang2022inner, brohan2023can, lin2023text2motion, wang2023describe},
and visual planning~\cite{kapelyukh2023dall, black2023zero, du2023video, du2024learning, ajay2024compositional, gao2024can}.
Among them, our work is most related to prior works that employ image or video prediction models to generate intermediate subgoal images
for the given language task~\cite{kapelyukh2023dall, black2023zero, du2023video, du2024learning, ajay2024compositional, gao2024can, zhou2024autonomous}.
These works use diffusion models to convert language instructions into visual subgoal plans,
which are then fed into low-level subgoal-conditioned policies to produce actions.
While sensible, this configuration leads to failures due to the misalignment of the generative models and the low-level policies that control the robot behavior,
as shown in our experiments (\cref{sec:experiments}).

\textbf{Rejection Sampling: }
One of our key ideas in this paper is based on rejection sampling, where we sample multiple subgoal proposals from an image or video prediction model and pick the best one based on a learned subgoal classifier.
The idea of test-time rejection sampling has been widely used in diverse areas of machine learning,
such as filtering-based action selection in offline reinforcement learning (RL)~\cite{fujimoto2019off, ghasemipour2021emaq, chen2022offline, hansen2023idql,nakamoto2024steering},
response verification in natural language processing~\cite{cobbe2021training, lightman2023let, hosseini2024v},
and planning and exploration in robotics~\cite{liu2022structdiffusion, brohan2023can, lin2023text2motion, huang2023grounded, ren2024explore}.
Previous works in robotics have proposed several ways to filter out infeasible plans generated by
pre-trained foundation models~\cite{liu2022structdiffusion, brohan2023can, lin2023text2motion, huang2023grounded, ren2023robots,myers2024policy}.
Unlike these works,
we focus on filtering visual subgoals instead of language plans~\cite{brohan2023can, huang2023grounded, ren2023robots},
and do not involve any planning procedures~\cite{lin2023text2motion} or structural knowledge~\cite{liu2022structdiffusion}.
While the subgoal classifier we train resembles the classifier from~\cite{nair2021lorel}, our classifier differs in two key ways. First, we use our classifier to filter out ``off-task'' subgoals, whereas the classifier in~\cite{nair2021lorel} is used as a reward function for training downstream policies. Second, the classifier from~\cite{nair2021lorel} is conditioned on the initial state $s_0$ and the current state $s$, whereas our classifier is conditioned on the current state $s$ and a generated subgoal $g$.

\textbf{Goal-Conditioned Policy Learning: }
Our method is broadly related to goal-conditioned policy learning~\cite{kaelbling1993learning, schaul2015universal, andrychowicz2017hindsight},
language-conditioned policy learning~\cite{tellex2020robots, stepputtis2020language,mees2022hulc,mees23hulc2,lynch2020language,hirose24lelan},
and hierarchical control~\cite{black2023zero, du2024learning, mandlekar2020iris, park2024hiql, sutton1999between, bacon2017option}.
Most prior works in hierarchical policy learning either train a high-level policy from scratch that produces subgoals or latent skills~\cite{schmidhuber1991learning, dayan1992feudal, kulkarni2016hierarchical, vezhnevets2017feudal, levy2017learning, nachum2018data, nachum2018near, gupta2019relay, ajay2020opal, lynch2020learning, rosete2023latent, zhang2020generating, pertsch2021accelerating, chane2021goal, park2024hiql}
or employ subgoal planning~\cite{savinov2018semi, eysenbach2019search, nair2019hierarchical, nasiriany2019planning, huang2019mapping, hoang2021successor, kim2021landmark, zhang2021world, shah2021rapid, fang2022planning, li2022hierarchical, kim2023imitating, rosete2023latent, fang2023generalization}.
Unlike these works, we do not train a high-level subgoal prediction model from scratch nor involve a potentially complex planning procedure.
Instead, we sample multiple potential subgoals from a pre-trained (or potentially fine-tuned) image or video prediction model and pick the best one based on a trained subgoal classifier.
Among hierarchical policy methods, perhaps the closest work to ours is IRIS~\cite{mandlekar2020iris},
which trains a conditional variational autoencoder to generate subgoal proposals and selects the best subgoal that maximizes the task value function.
While conceptually similar, our method differs from IRIS in that we do not assume access to a reward function in order to train a value function. 
Our classifier is trained on trajectories consisting only of images and language descriptions.

\textbf{Diffusion Model Guidance: }
The generative models we consider in our paper~\cite{brooks_instrucpix2pix, xing2023dynamicrafter} are diffusion-based models trained using classifier-free guidance (CfG)~\cite{ho2022classifier}. Although we use a large value for the language-prompt guidance parameter at inference in our experiments, we find that producing ``off-task'' subgoals is still a common failure mode that is not solved by increasing this parameter alone.

Classifier guidance~\cite{sohl2015deep, song2020score, dhariwal2021diffusion} is also a plausible alternative to rejection sampling, but there are some practical challenges in training a subgoal classifier for this purpose. First, the diffusion models we consider use latent diffusion~\cite{rombach2022high}, and therefore would require training the subgoal classifier to operate in the latent space of the diffusion model. Second, the subgoal classifier would need to be trained on noised data in order to guide the diffusion denoising process of the generative model.
Nevertheless, classifier guidance is a potentially appealing direction for future work.

\section{Preliminaries}

We consider the same problem setting as~\cite{black2023zero}, where the goal is for a robot to perform a task described by some previously unseen language command $l$. To do this, we consider the same three dataset categories as in~\cite{black2023zero}: (1) language-labeled video clips $\mathcal{D}_l$ which contain no robot actions; (2) language-labeled robot data $\mathcal{D}_{l, a}$ that includes both language labels and robot actions; (3) unlabeled robot data that only includes actions $\mathcal{D}_a$. The dataset $\mathcal{D}_{l, a}$ consists of a set of trajectory and task language pairs, $\{(\tau^n, l^n)\}_{n=1}^{N}$, and a trajectory contains a sequence of state, $s_{t}^{n} \in \mathcal{S}$, and action, $a_{t}^{n} \in \mathcal{A}$, pairs, $\tau^{n} = (s_{0}^{n}, a_{0}^{n}, s_{1}^{n}, a_{1}^{n}, \ldots)$.
Given these datasets, we assume access to two learned modules:
\begin{enumerate}
    \item \textbf{a subgoal generation module} from which we can sample multiple possible future subgoals. This can be trained on $\mathcal{D}_l$ and $\mathcal{D}_{l, a}$.
    \item \textbf{a low-level goal-reaching policy} that chooses actions to reach generated subgoals. This can be trained on $\mathcal{D}_a$ and/or $\mathcal{D}_{l, a}$.
\end{enumerate}
Our contribution is a set of approaches to robustify the interface between these two modules.

While \algname can be applied to any hierarchical imitation learning method consisting of the two components mentioned above, in this work we apply \algname to two specific algorithms: (1) UniPi~\cite{du2024learning}, in which a high-level model generates a subgoal video, and a low-level inverse-dynamics model predicts the actions needed to ``connect'' the images in the video, and (2) SuSIE~\cite{black2023zero}, in which a high-level model generates a subgoal image by ``editing'' the current image observation, and a  goal-conditioned policy predicts actions to achieve the subgoal image. We define subgoals, $g \in \mathcal{G}$, as video or image samples from the high-level models used in these algorithms.

\section{\algname}

Many modern hierarchical policy methods aim to improve generalization by using language-conditioned image or video models to generate intermediate subgoal images for a given task. The interface between these image or video models and the low-level policies that choose actions to reach generated subgoals is a major performance bottleneck for these hierarchical policy methods. \algname improves the robustness of this interface (see~\cref{fig:method_teaser}). In~\cref{subsec:method_subgoal_filtering}, we propose a simple method to filter subgoals that do not progress towards completing the task specified by language instruction $l$. Then, in~\cref{subsec:method_image_aug}, we describe a simple yet non-obvious data augmentation practice to robustify the low-level policy and our subgoal classifier to harmful visual artifacts in the generated subgoals. 
We note that the two components of \algname work together synergistically: when applied together, the resulting performance improvement is larger than the sum of improvements that results from applying each component individually (see \cref{sec:experiments}).

\subsection{Subgoal Filtering}
\label{subsec:method_subgoal_filtering}
The image and video generative models we consider are first pre-trained on general Internet-scale image and video data, and then fine-tuned on a modest amount of robot data. Despite being fine-tuned on robot data, a common failure mode we observe across different models is that, over the course of executing a task,
the model begins to go ``off-task,'' meaning that it starts generating subgoals that are consistent with the current image observation but that do not progress towards completing the language instruction $l$.
We hypothesize that this is due to the distribution shift between the Internet data these image or video prediction models are pre-trained on and the robot data they are fine-tuned on. 

To address this challenge, we train a subgoal classifier $f_\theta(s, g, l)$ on $\mathcal{D}_{l}$ and/or $\mathcal{D}_{l,a}$ that predicts the probability that the transition between the current image observation $s$ and the next subgoal $g$ makes progress towards completing language instruction $l$. Note that although we train the subgoal classifier on 
robot data
in our experiments, action labels are not used in the training of the classifier, and the subgoal classifier can be trained on action-free data, including large, non-robotics Internet video datasets. During training, we sample positive examples of state-goal transitions for $l$ from the set of trajectories that successfully complete the instruction. We construct negative examples in the following three ways:
\begin{enumerate}
    \item \textbf{Wrong Instruction:} $(s, g, l')$ where $l'$ is sampled from a different transition than $s$ and $g$.
    \item \textbf{Wrong Goal Image:} $(s, g', l)$ where $g'$ is sampled from a different transition than $s$ and $l$.
    \item \textbf{Reverse Direction}: $(g, s, l)$, where the order of the current image observation and the subgoal image have been switched. This is important for learning whether a candidate goal image is making temporal progress towards completing the language instruction. 
\end{enumerate}
We refer to this dataset of negative examples constructed from $\mathcal{D}_{l,a}$ as $\mathcal{D}_{l,a}^-$.
We then train the subgoal classifier by minimizing the binary cross entropy loss between the positive examples and the constructed negative examples (see ~\cref{appendix:classifier_training} for additional training details):

\begin{equation}
    \begin{split}
        \mathcal{J}(\theta) &= \expectation_{(s, g, l) \sim \mathcal{D}_{l,a}} \left[ \log \left( f_\theta(s, g, l) \right) \right] \\
        &\quad + \expectation_{(s^-, g^-, l^-) \sim \mathcal{D}_{l,a}^-} \left[ \log \left( 1 - f_\theta(s^-, g^-, l^-) \right) \right].
    \end{split}
\end{equation}

 Given a set of $K$ subgoals predicted by the image or video model, \algname uses the classifier to select the subgoal with the highest progress probability and passes that subgoal to the low-level policy for conditioning.

\subsection{Image Augmentation De-Synchronization}
\label{subsec:method_image_aug}

While the method proposed in~\cref{subsec:method_subgoal_filtering} increases robustness to predicted subgoals that do not make task progress, generated subgoals can also contain visual artifacts that degrade the performance of both the low-level control policy and the subgoal classifier. 
This performance degradation results from the distribution shift between the subgoal images seen by the policy during training, which come from the robot dataset, and the subgoal images seen during inference, which come from the generative model. 
Ideally, the low-level policy and subgoal classifier would be trained on the same distribution of \textit{generated} subgoal images that they will see at inference time. 
However, due to the high degree of variance in sampling images from a generative model, there is not a clear way to obtain generated subgoal images that match the actual future states reached in trajectories in the training data. 
To address this issue, we identify a simple yet non-obvious data augmentation practice to train the low-level policy and subgoal classifier on goals from the robot dataset while also robustifying them to visual artifacts in generated subgoals.

Applying image augmentation procedures such as random cropping or color jitter during training is a standard approach in image-based robot learning methods~\cite{domain_rand} to improve the robustness of learned models to distribution shifts between their training and evaluation domains. More formally, let $\phi$ be the set of image augmentation parameters to be randomly sampled from space $\Phi$, $p_\Phi(\cdot)$ be some probability distribution over $\Phi$, and let $\hat{\phi} \sim p_\Phi(\cdot)$ be some realization of augmentations sampled from $p_\Phi(\cdot)$. Typically, for each training sample, a different value $\hat{\phi}$ is applied during training to make a model robust to any augmentation in the space $\Phi$. 

For both the low-level goal-conditioned policy and the subgoal classifier, each training sample includes two images: the current state $s$ and the corresponding goal $g$. Standard practice is to sample augmentation parameters $\hat{\phi}$ and apply them to all images in a given training sample~\cite{black2023zero, zheng2023stabilizing}, which corresponds to applying the same $\hat{\phi}$ to both $s$ and $g$. In a non-hierarchical policy setting, this makes sense, because at inference time $s$ and $g$ will both be sampled from the camera observations of the current environment instantiation. 
However, when using an image or video prediction model for subgoal generation, at inference time the low-level policy and subgoal classifier will see states from the camera observations, but the goals will be generated by the image or video prediction model. 
There will often be differences in the visual artifacts between a camera observation $s$ and the corresponding generated subgoal image $g$, 
such as differences in color, contrast, blurriness, and the shapes of objects, which can degrade the performance of low-level policies and subgoal classifiers. 

To encourage robustness to this distribution shift, we sample separate augmentation parameters for $s$ and $g$, denoted by $\hat{\phi_s}$ and $\hat{\phi_g}$ (i.e., we de-synchronize the image augmentations applied to $s$ and $g$). Random cropping, brightness shifts, contrast shifts, saturation shifts, and hue shifts comprise our space of augmentations (see~\cref{appendix:image_augmentations} for details). 
Concretely, for each $s$ and $g$ pair sampled during training, a different random crop, brightness, contrast, saturation, and hue shift are applied to $s$ than are applied to $g$. 
This forces the low-level policy and the subgoal classifier to learn to make accurate predictions on $(s, g)$ pairs that have differences in visual artifacts.

While image augmentations are ubiquitous in robot learning methods, our experiments show that the standard way of applying image augmentations for goal-conditioned policies and classifiers is deficient for the hierarchical policy methods that we consider.
We also note that augmentation de-synchronization is applied not only to the policy, but also to the subgoal classifier (\cref{subsec:method_subgoal_filtering}), which has a significant impact on overall performance (\cref{sec:experiments}).

\section{Experiments}\label{sec:experiments}

\begin{figure}[t]
    \centering
    \vspace{2mm}
    \includegraphics[width=0.99\linewidth]{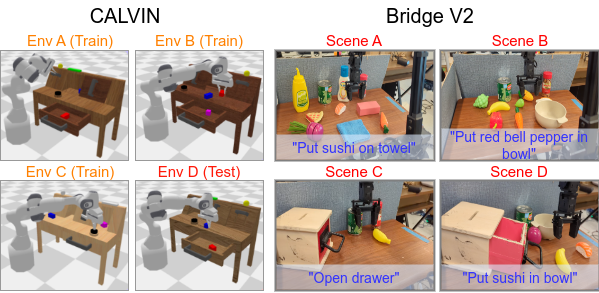}
    \vspace{-6mm}
    \captionsetup{font=small}
    \caption{
        \textbf{Experimental Domains. } \underline{Simulation Environments (Left):} Train/test environments in the CALVIN simulation benchmark. The environments each have different table textures, furniture positions, and initial configurations of the colored blocks. Each environment contains 34 tasks, each with an associated language instruction. To test zero-shot generalization, environment D is held out for evaluation. \underline{Physical Environments (Right):} We consider four test scenes in the Bridge~V2 robot platform with four total language instructions. 
        To test zero-shot generalization, these test scenes contain novel objects, language commands, and object configurations not seen in the training data.
    }
    \label{fig:exp_domains}
    \vspace{-4mm}
\end{figure}

We study the degree to which \algname improves existing hierarchical imitation learning algorithms across a number of tasks in simulation and physical experiments that assess zero-shot generalization. 
On the CALVIN~\cite{mees_calvin} simulation benchmark, we find that applying \algname to two different hierarchical methods that leverage generative subgoals yields an average increase in relative performance of 27\%. In experiments on a physical robot, \algname increases the relative success rate of a SOTA generative hierarchical imitation learning method~\cite{black2023zero} by 23\%. The improvement in average relative performance yielded by \algname across both simulated and real experiments is 25\%.
We analyze the influence of each component of \algname on task performance and also perform extensive qualitative analysis in~\cref{appendix:qualitative_analysis}.

\subsection{Experimental Domains}
\label{subsec:exp_domains}
We evaluate our method on the CALVIN~\cite{mees_calvin} simulation benchmark and the Bridge V2~\cite{walke_bridgev2} physical experiment setup with a WidowX250 robot.

\begin{figure*}[t]
    \centering
    \vspace{2mm}
    \includegraphics[width=0.95\linewidth]{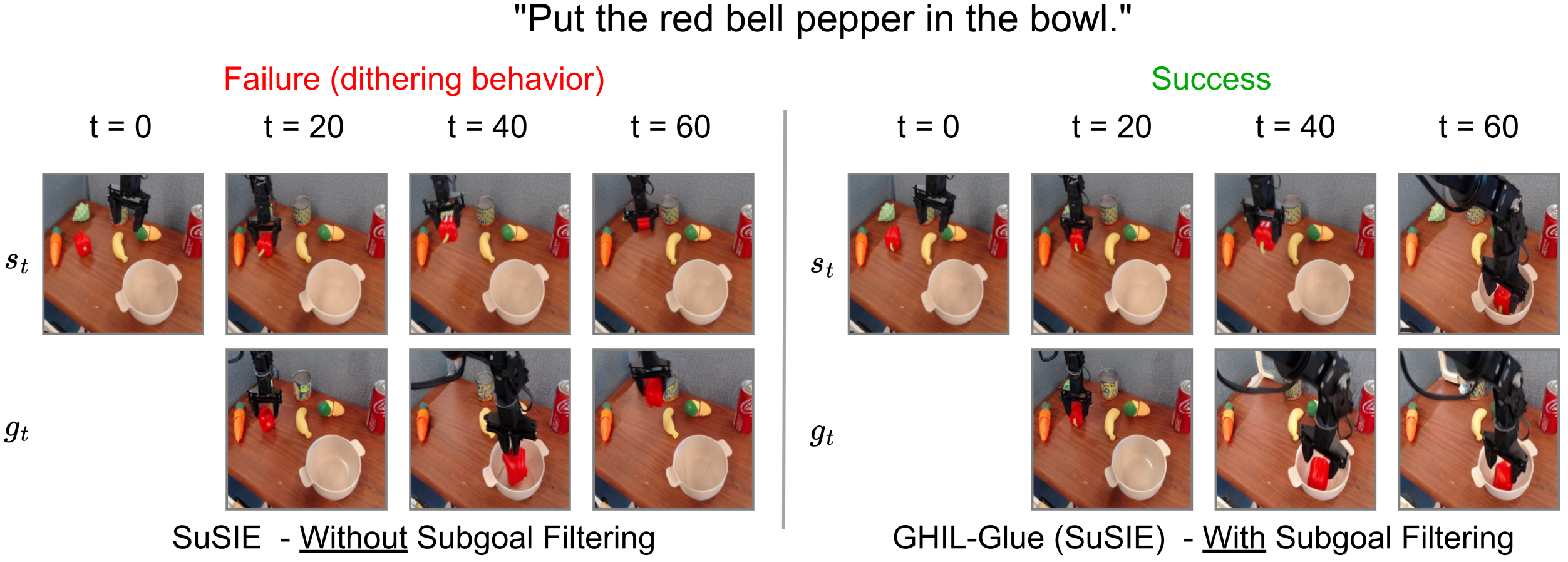}
    \vspace{-1mm}
    \captionsetup{font=small}
    \caption{
        \textbf{\algname Subgoal Filtering.} We visualize policy rollouts of SuSIE without subgoal filtering vs. \algname SuSIE with subgoal filtering. We show the states reached every 20 timesteps (top row) and the corresponding predicted subgoals (bottom row). Without subgoal filtering, the subgoal at $t=60$ is not consistent with making progress towards placing the pepper in the bowl, causing the robot to dither and drop the pepper. When subgoal filtering is used, the selected subgoals make iterative progress towards a successful task completion.
    }
    \label{fig:qualitative}
    \vspace{-6mm}
\end{figure*}

\textbf{Simulation Experiment Setup: }
Simulation experiments are performed in the CALVIN~\cite{mees_calvin} benchmark, which focuses on long-horizon language-conditioned robot manipulation. We follow the same protocol as in~\cite{black2023zero}, and train on data from three environments (A, B, and C) and test policies on a fully unseen environment (D). Each environment contains a Franka Emika Panda robot arm that is placed in front of a desk with a variety of objects and is associated with 34 possible tasks (\cref{fig:exp_domains}).
The held-out environment (D) contains unseen desk and object colors, object and furniture positions, and object shapes. The corresponding language instructions are similarly held out.

\textbf{Physical Experiment Setup: }
For physical experiments, we use the same datasets as in~\cite{black2023zero} for training both the high-level image prediction model and the low-level goal-conditioned policy. The Bridge V2 dataset contains 45K language-annotated trajectories, which are used for the language-labeled robot dataset $\mathcal{D}_{l, a}$. The remaining 15K trajectories are used for the action-only dataset $\mathcal{D}_a$.
As in~\cite{black2023zero}, we use a filtered version of the Something-Something V2 dataset~\cite{goyal_something_something} with the same filtering scheme as in~\cite{black2023zero} (resulting in 75K video clips) as our video-only dataset $\mathcal{D}_l$. 

We test our policies on four tasks on four different cluttered table top scenes (\cref{fig:exp_domains}) on the Bridge V2 physical robot platform.  
These environments require generalizing to novel scenes, with novel objects, and with novel language commands that are not seen in the Bridge V2 dataset. 

\subsection{Comparison Algorithms}
\label{subsec:exp_algorithms}
To evaluate \algname's performance, we study the impact of applying it to two SOTA hierarchical imitation learning algorithms: SuSIE~\cite{black2023zero} and UniPi~\cite{du2024learning}. To evaluate the importance of hierarchy more generally, we also compare \algname to a flat language-conditioned diffusion policy (LCBC Diffusion Policy). Finally, we consider ablations where we separately study the impact of each of our proposed contributions: subgoal filtering (\cref{subsec:method_subgoal_filtering}) and de-synchronizing augmentations (\cref{subsec:method_image_aug}). For physical experiments, we additionally consider a comparison to OpenVLA~\cite{kim24openvla},
which is trained on the Open X-Embodiment dataset~\cite{open_x_embodiment_rt_x_2023} (which includes the Bridge V2 dataset). 
\begin{enumerate}
    \item \textbf{LCBC Diffusion Policy: } Low-level language-conditioned behavior cloning diffusion policy~\cite{chi2023diffusion} trained only on robot trajectories with language annotations. We use the same implementation as in~\cite{black2023zero}.
    \item \textbf{OpenVLA~\cite{brohan2023rt}: } A SOTA language-conditioned vision-language-action model (VLA) trained on the Open X-Embodiment dataset~\cite{open_x_embodiment_rt_x_2023} (which includes the entirety of the Bridge V2 dataset).
    \item \textbf{SuSIE~\cite{black2023zero}: } A method which fine-tunes InstructPix2Pix~\cite{brooks_instrucpix2pix}, an image-editing diffusion model, to generate subgoal images given the current image observation. Low-level control is performed using a goal-conditioned policy. For SuSIE and all methods that build on it, we predict subgoals 20 steps in the future as in the original paper.
    \item \textbf{UniPi~\cite{du2024learning}: } A method which fine-tunes a language-conditioned video prediction model on robot data and then uses an inverse dynamics model for low-level goal reaching. 
    For UniPi and all methods that build on it, we predict video sequences of 16 frames. 
    As the original UniPi model is not publicly available, we re-implement UniPi by fine-tuning the video model from~\cite{xing2023dynamicrafter}.
    \item \textbf{\algname (SuSIE / UniPi): } \algname applied on top of either SuSIE or UniPi. For all experiments we implement the subgoal filtering step by sampling four to eight subgoals from the high-level video prediction model and selecting amongst them
    (see~\cref{appendix:ablations} for details). 
    We directly filter the subgoal images generated by the SuSIE model. We filter the video sequences generated by the UniPi model based on the final frame of each sequence. 
    \item \textbf{\algname (SuSIE / UniPi) - Subgoal Filtering Only: } \algname applied to SuSIE or UniPi using subgoal filtering but without augmentation de-synchronization.
    \item \textbf{\algname (SuSIE / UniPi) - Aug De-sync Only: } \algname applied to SuSIE or UniPi using augmentation de-synchronization but without subgoal filtering.
\end{enumerate}

\begin{table*}[h!]
\centering
\vspace{2mm}
\begin{tabular}{clcccccc}
\toprule
 & \multicolumn{1}{c}{} & \multicolumn{5}{c}{Tasks completed in a row} \\ \cmidrule(lr){3-8}
 & \multicolumn{1}{c}{\multirow{-2}{*}{Method}}  & 1  & 2  & 3  & 4  & 5  & Avg. Len.\\ \midrule
 & LCBC Diffusion Policy & 68.5\% & 43.0\% & 22.5\% & 11.0\% & 6.8\% & 1.52 \\
& SuSIE~\cite{black2023zero} & 89.8\% & 75.0\% & 57.5\% & 41.8\% & 29.8\% & 2.94 \\ %
& \algname (SuSIE) - Aug De-sync Only & 95.2\% & 84.0\% & 69.5\% & 56.0\% & 46.2\% & 3.51 \\
& \algname (SuSIE) - Subgoal Filtering Only & 88.5\% & 75.5\% & 56.2\% & 43.0\% & 32.5\% & 2.96 \\
& \algname (SuSIE) & \cellcolor[HTML]{B7E1CD}\textbf{95.2\%} & \cellcolor[HTML]{B7E1CD}\textbf{88.5\%} & \cellcolor[HTML]{B7E1CD}\textbf{73.2\%} & \cellcolor[HTML]{B7E1CD}\textbf{62.5\%} & \cellcolor[HTML]{B7E1CD}\textbf{49.8\%} & \cellcolor[HTML]{B7E1CD}\textbf{3.69} \\
& UniPi~\cite{du2024learning} & 56.8\% & 28.3\% & 12.0\% & 3.5\% & 1.5\% & 1.02 \\
& \algname (UniPi) - Aug De-sync Only & 60.2\% & 29.5\% & 12.5\% & 5.5\% & 1.8\% & 1.1 \\
& \algname (UniPi) - Subgoal Filtering Only & 69.5\% & 40.0\% & 15.8\% & 6.5\% & 4.2\% & 1.36 \\
& \algname (UniPi) & \cellcolor[HTML]{B7E1CD}75.2\% & \cellcolor[HTML]{B7E1CD}44.8\% & \cellcolor[HTML]{B7E1CD}19.7\% & \cellcolor[HTML]{B7E1CD}11.2\% & \cellcolor[HTML]{B7E1CD}5.5\% & \cellcolor[HTML]{B7E1CD}1.56 \\
\bottomrule
\end{tabular}
\vspace{-1mm}
\captionsetup{font=small}
\caption{\textbf{CALVIN: Simulation Results. } Success rates on the  validation tasks from the held-out D environment of the CALVIN zero-shot generalization challenge averaged across 4 random seeds. Applying \algname to SuSIE and UniPi significantly improves performance over their respective base methods. \algname (SuSIE) significantly outperforms all other methods, 
achieving a
new state-of-the-art on the CALVIN benchmark for policies using observations from a single RGB camera. }
\label{table:calvin_susie_exps}
\end{table*}

\begin{table*}[h!]
\centering
\begin{tabular}{llccc}
\toprule
& \multicolumn{1}{l}{\multirow{1}{*}{Task}}  & OpenVLA~\cite{kim24openvla}  & SuSIE~\cite{black2023zero} & \algname(SuSIE)\\ \midrule
Scene A & Put Sushi On Towel & 22/30 & 19/30 & \textbf{28/30}\\
Scene B & Put Red Bell Pepper in Bowl & 14/30 & 12/30 & \textbf{16/30}\\
Scene C & Open Drawer & \textbf{23/30} & 19/30 & 22/30\\
Scene D & Put Sushi in Bowl & 15/30 & 15/30 & \textbf{18/30}\\
\bottomrule
\end{tabular}
\vspace{-1mm}
\captionsetup{font=small}
\caption{\textbf{Bridge V2 Physical Experiments Results. } Success rates across four tasks on four physical robot scenes (pictured in~\cref{fig:exp_domains}) that test zero-shot generalization to novel objects, novel language commands, and novel scene configurations. \algname applied to SuSIE outperforms SuSIE across all tasks and outperforms OpenVLA on 3 out of 4 tasks.
}
\label{table:bridge_exps}
\vspace{-6mm}
\end{table*}

\subsection{Experimental Results}
\label{subsec:exp_results}

\textbf{Simulation Experiments: }We present results on the CALVIN benchmark in~\cref{table:calvin_susie_exps}. 
Applying \algname yields significant performance increases for SuSIE and UniPi, increasing the average successful task sequence length from \textbf{2.94} to \textbf{3.69} for SuSIE and from \textbf{1.02} to \textbf{1.56} for UniPi. 
This constitutes an average increase in relative successful sequence length of 27\%.
\textbf{\algname (SuSIE) achieves a new SOTA on CALVIN} for policies that use observations from a single RGB camera. 

The two components of \algname (subgoal filtering and image augmentation de-synchronization) improve performance when applied individually, but, when applied together, these components build on each other, leading to a performance increase greater than the sum of the individual benefits.
Specifically, for SuSIE, image augmentation de-synchronization and subgoal filtering individually yield increases in sequence length of 0.56 and 0.02 respectively, whereas when applied together they yield an increase of 0.75. Similarly, for UniPi, the individual improvements yield increases in sequence length of 0.08 and 0.34 respectively, compared to an increase of 0.54 when applied together.

When applied alone, image augmentation de-synchronization increases the average successful task sequence length from 2.94 to 3.51 for SuSIE and from 1.02 to 1.1 for UniPi. 
We hypothesize that augmentation de-synchronization improves performance a large amount with SuSIE because its low-level policy is conditioned on a camera observation image $s$ from the environment and a subgoal image $g$ generated by the image model. 
When generalizing to the held-out test environment D, the SuSIE image model generates subgoal images with visual discrepancies from the camera observation images.
In contrast, the UniPi video model predicts a sequence of frames as opposed to a single subgoal image. The UniPi low-level policy functions as an inverse dynamics model, choosing actions to link between the frames of the generated subgoal video, and is therefore conditioned on an $s$ and $g$ that both come from the predicted subgoal video.

When applied alone, subgoal filtering has a small effect on SuSIE, while on UniPi it increases the average successful task sequence length from 1.02 to 1.36. 
This suggests that unless the SuSIE low-level policy is made robust to visual artifacts in generated subgoals, simply selecting the most task relevant subgoal is insufficient to improve performance.
As discussed previously, the SuSIE low-level policy is more sensitive to visual artifacts in generated subgoals than is the UniPi inverse dynamics model.

\textbf{Physical Experiments: }We present results (\cref{table:bridge_exps}) comparing \algname (SuSIE) to OpenVLA and SuSIE across four environments on the Bridge V2 robot platform that require interacting with a number of objects on a cluttered table (\cref{fig:exp_domains}). These environments require generalizing to novel scenes, with novel objects, and with novel language commands that are not seen in the Bridge V2 dataset. 
\algname applied to SuSIE outperforms SuSIE across all tasks and increases the overall success rate from 54\% to 70\%, yielding a 23\% relative increase in success rate.
\algname (SuSIE) also outperforms OpenVLA, a 7-billion parameter SOTA VLA, on 3 out of 4 tasks. 
Significantly, the baseline SuSIE implementation does not outperform OpenVLA on a single task, whereas \algname (SuSIE) outperforms OpenVLA on 3 out of 4 tasks, demonstrating that hierarchical goal conditioned architectures with well-tuned interfaces between the high and low-level policies can outperform SOTA VLA methods on zero-shot generalization tasks.
See~\cref{appendix:qualitative_analysis} for qualitative examples of success and failure cases of \algname in physical experiments, and for examples of generated subgoals for a subset of the tasks in addition to their scores under our subgoal filtering method.

\section{Conclusion}

We present \algname, a method for better aligning image and video prediction models and low-level control policies for hierarchical imitation learning. Our key insight is that while image and video foundation models can generate highly realistic subgoals for goal-conditioned policy learning, when generalizing to novel environments, the generated images are prone to containing visual artifacts and can be inconsistent with the task the robot is commanded to perform. \algname provides two simple ideas to address these challenges, leading to a significant increase in zero-shot generalization performance over prior work both in the CALVIN simulation benchmark and in physical experiments.

One exciting avenue for future work would be to explore training image or video prediction models for subgoal generation on a broader distribution of robot data, such as the data available in the Open-X embodiment dataset~\cite{open_x_embodiment_rt_x_2023}.
Another interesting direction would be to filter subgoals based on the capability of the low-level policy to actually achieve them, for example, by training a goal-conditioned value function for the low-level policy and using it to evaluate subgoal feasibility.
Finally, while we trained the subgoal classifiers on robot datasets, in principle these could be trained in the same way on much larger, non-robotics video datasets in order to improve generalization.

\section{Acknowledgments}

We thank Kevin Black, Pranav Atreya, and Mitsuhiko Nakamoto for their guidance with SuSIE~\cite{black2023zero}. Authors from University of California, Berkeley and Princeton University were partially supported by funding from Toyota Research Institute (TRI).

\bibliographystyle{IEEEtran}
\bibliography{refs}

\clearpage
\onecolumn

\appendix
\label{appendix}

\subsection{Classifier Training}
\label{appendix:classifier_training}

\textbf{Training objective:}
The classifier is trained using binary cross-entropy loss: 

\begin{equation}
    \begin{split}
        \mathcal{J}(\theta) &= \expectation_{(s, g, l) \sim \mathcal{D}_{l,a}} \left[ \log \left( f_\theta(s, g, l) \right) \right] \\
        &\quad + \expectation_{(s^-, g^-, l^-) \sim \mathcal{D}_{l,a}^-} \left[ \log \left( 1 - f_\theta(s^-, g^-, l^-) \right) \right].
    \end{split}
\end{equation}

where $D_{l,a}$ is the language-annotated dataset that consists of trajectory and language task pairs, and $N$ is a function for generating negative examples from the dataset. 
Given a dataset $D_{l,a}$, $N$ generates negatives from  $D_{l,a}$ in the following ways:  
\begin{enumerate}
    \item \textbf{Wrong Instruction:} $(s, g, l')$ where $l'$ is sampled from a different transition than $s$ and $g$.
    \item \textbf{Wrong Goal Image:} $(s, g', l)$ where $g'$ is sampled from a different transition than $s$ and $l$.
    \item \textbf{Reverse Direction}: $(g, s, l)$, where the order of the current image observation and the subgoal image have been switched. 
\end{enumerate}

Across all our experiments, we sample $50\%$ of each training batch to be positive examples and $50\%$ of each training batch to be negative examples. Of the negative examples, $40\%$ are ``wrong instruction'', $40\%$ are ``reverse direction'', and $20\%$ are ``wrong goal image''.

\textbf{Goal sampling:} In a given training tuple $(s_t, g, l)$, $g$ is sampled by taking the goal image from the $s_{t+k}$, where $k$ is a uniformly sampled integer from 16 to 24.

\textbf{Network architecture and training hyperparameters:}
The classifier network architecture consists of a ResNet-34 encoder from \cite{walke_bridgev2}, followed by a two-layer MLP with layers of dimension 256. Separate encoders are used to encode the image observations and the goal images (parameters are not shared between the two). Both of these encoders use FiLM conditioning~\cite{perez2018film} after each residual block to condition on the language instruction.
Classifier networks are trained using a learning rate of $3\times 10^{-4}$ and a batch size of $256$ for $100,000$ gradient steps. A dropout rate of $0.1$ is used.

\subsection{Image Augmentations}
\label{appendix:image_augmentations}

\begin{wrapfigure}[10]{R}{0.5\linewidth}
\vspace{-6mm}
    \centering
    \includegraphics[width = \linewidth]{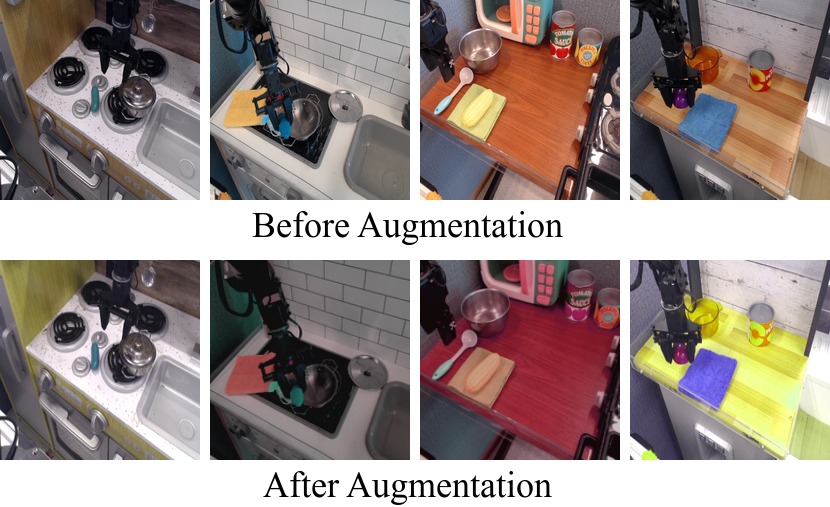}
    \captionsetup{font=small}
    \caption{
        \textbf{Image augmentation examples} Examples of images from the Bridge dataset before and after having the image augmentations applied to them that are used during policy and classifier training. 
    }
    \label{fig:augmentation_examples}

\end{wrapfigure}

During training of low-level policy networks and classifier networks, we apply the following augmentations to the image observations and the goal images, in the following order:
\begin{enumerate}
    \item Random Resized Crop: 
        \begin{itemize}
            \item scale: $(0.8, 1.0)$
            \item ratio:$(0.9, 1.1)$
        \end{itemize}
    
    \item Random Brightness Shift: 
        \begin{itemize}
            \item shift ratio: $0.2$
        \end{itemize}

    \item Random Contrast: 
        \begin{itemize}
            \item Contrast range: $(0.8, 1.2)$
        \end{itemize}

    \item Random Saturation: 
        \begin{itemize}
            \item Saturation range: $(0.8, 1.2)$
        \end{itemize}

    \item Random Hue: 
        \begin{itemize}
            \item shift ratio: $0.1$
        \end{itemize}
        
\end{enumerate}

\Cref{fig:augmentation_examples}  visualizes examples from the Bridge dataset before and after augmentations are applied.

\subsection{Qualitative Analysis}
\label{appendix:qualitative_analysis}

\subsubsection{Classifier rankings}

We show examples of how the classifier network ranks generated goal images on tasks from Scene~D of our physical experimental domain. Figures~\ref{fig:classifier_ranking_examples3}, \ref{fig:classifier_ranking_examples4}, \ref{fig:classifier_ranking_examples5} show examples of the classifier correctly ranking the generated goal images (highly ranked images correspond to making progress towards correctly completing the language instruction), while \cref{fig:classifier_ranking_examples2} shows an example of the classifier erroneously giving high rankings to goal images that do not make progress towards completing the language instruction. 
Note that while the classifier scores can be close across various goal images, so long as the relative ranking of the generated goal images is correct, then incorrect subgoal images will be rejected and correct subgoal images will be passed to the low-level policy.

\subsubsection{Trajectory Visualizations}

\begin{figure}[h!]
    \centering
    \captionsetup{font=small}
    \caption{
            \textbf{\algname (SuSIE) Trajectory Visualization} 
            Visualization of a rollout of \algname (SuSIE) on Scene D in the physical experiments set up. The top row shows the current image observation at every timestep at which the video prediction model is queried. The second and third rows show the highest and lowest ranked generated subgoal images out of the 8 generated subgoal images, as ranked by the classifier. Note that during \algname inference, only the first-ranked subgoal is passed to the low-level policy.
        }
        
    \begin{subfigure}[b]{\linewidth}
        \centering
        \includegraphics[width=0.825\linewidth]{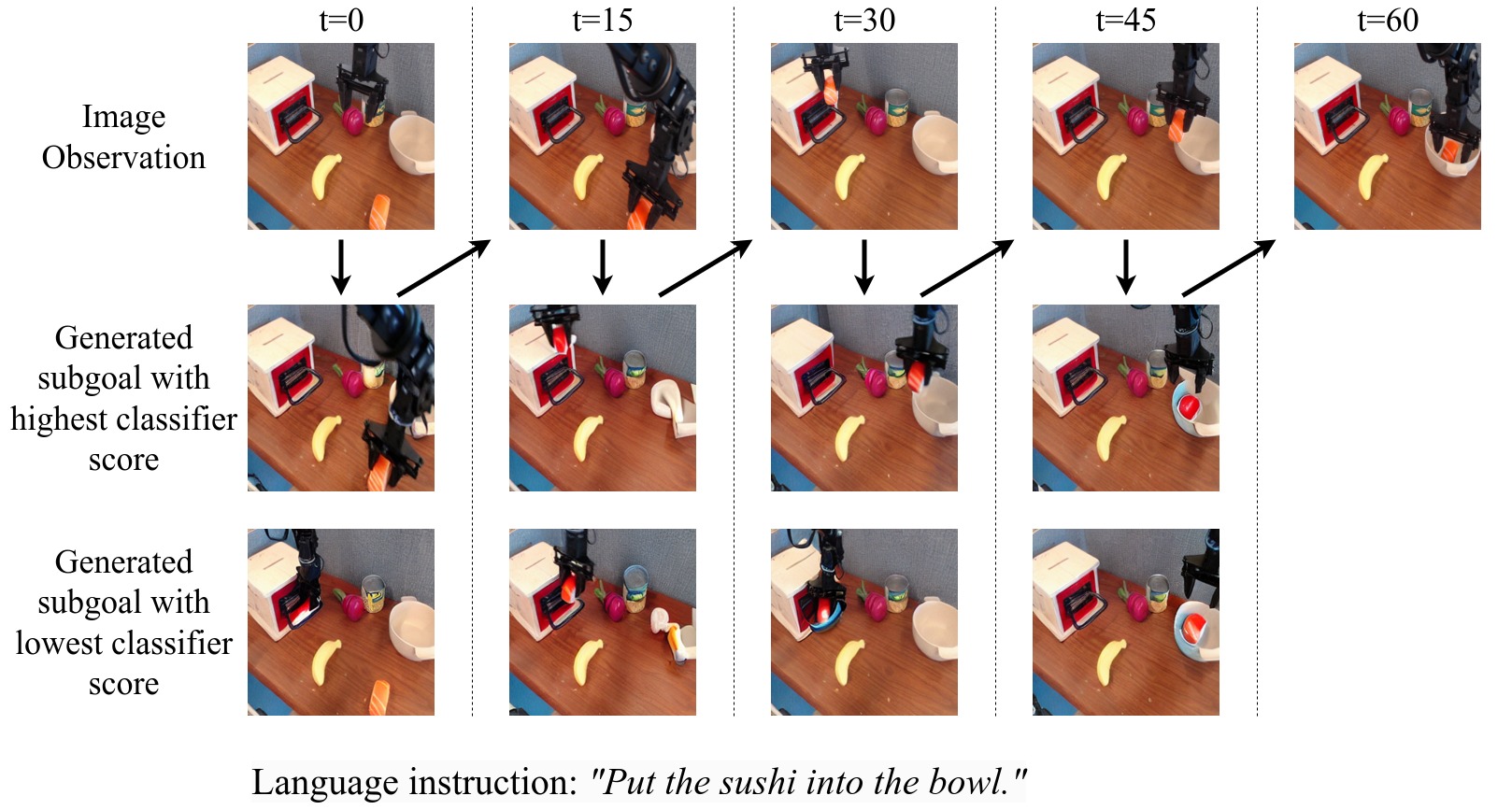}
        \captionsetup{font=small}
        \caption{
            \textbf{``Put the sushi into the bowl.''}
            This rollout shows two examples of the classifier filtering preventing the policy from going off-task: at $t=0$, the lowest ranked generated subgoal shows the gripper grasping the drawer handle instead of moving to grasp the sushi; at $t=30$, the lowest ranked generated subgoal shows the gripper moving towards the drawer handle instead of towards placing the sushi into the bowl. Note the hallucinated objects and artifacts visible in the goal images at $t=15, 30, 45$. Augmentation de-synchronization helps to make the low-level policy and classifier robust to hallucinated artifacts such as these.
        }
        \label{fig:traj_success}
    \end{subfigure}
    \vspace{2mm}
    
    \begin{subfigure}[b]{\linewidth}
        \centering
        \includegraphics[width=0.825 \linewidth]{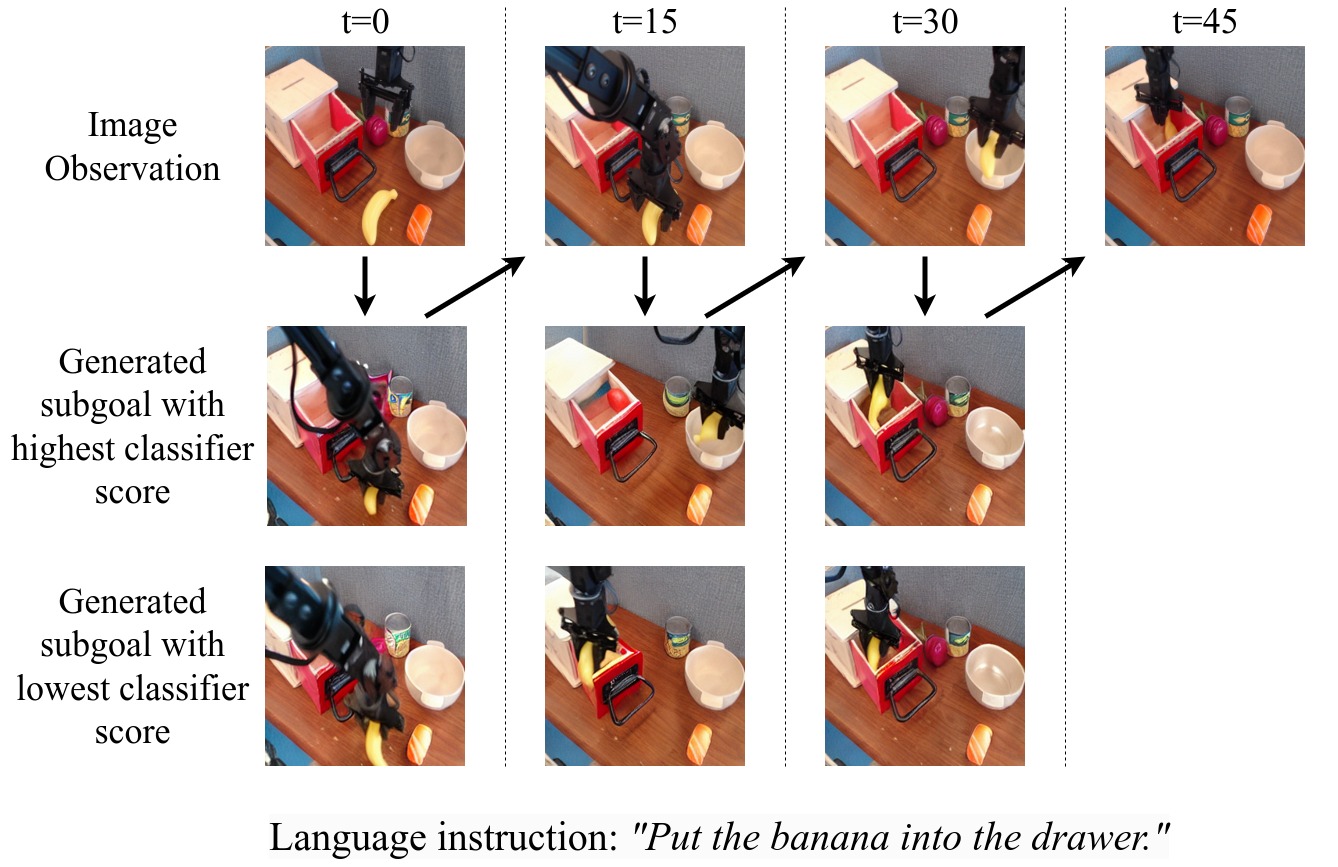}
        \captionsetup{font=small}
        \caption{
            \textbf{``Put the banana into the drawer.''} 
            In this rollout, classifier filtering fails and causes a near-miss. At $t=15$, the classifier ranks a subgoal image highest that shows the robot placing the banana into the bowl instead of the drawer. However, at $t=30$, when the robot reaches the state specified by this subgoal image, the subsequent generated subgoals all show the robot correctly placing the banana into the drawer. Although, as in this example, the classifier network can occasionally rank incorrect subgoal images higher than correct subgoal images, such errors occur infrequently as \algname (SuSIE/UniPI) outperforms base-SuSIE/UniPi across all of our physical and simulated experiments.
        }
        \label{fig:traj_near_miss}
    \end{subfigure}
        \label{fig:trajectory_visualizations}
\end{figure}

We show examples of rollouts of \algname (SuSIE) on our physical experiment set up. 
These examples showcase when \algname successfully filters out off-task subgoal images (\Cref{fig:traj_success}), 
as well as an instance of when \algname nearly causes a failure (\Cref{fig:traj_near_miss}).

\begin{figure}[h!]
    \centering
    \captionsetup{font=small}
    \caption{
            \textbf{Classifier ranking examples} 
            Examples of the classifier network rankings on 8 generated candidate subgoals given an observation from Scene D of the physical experiments and a language instruction. Note that during \algname inference, only the first-ranked subgoal is passed to the low-level policy.
        }
    \label{fig:classifier_ranking_examples}
    
    \begin{subfigure}[b]{\linewidth}
        \centering
        \includegraphics[width=\linewidth]{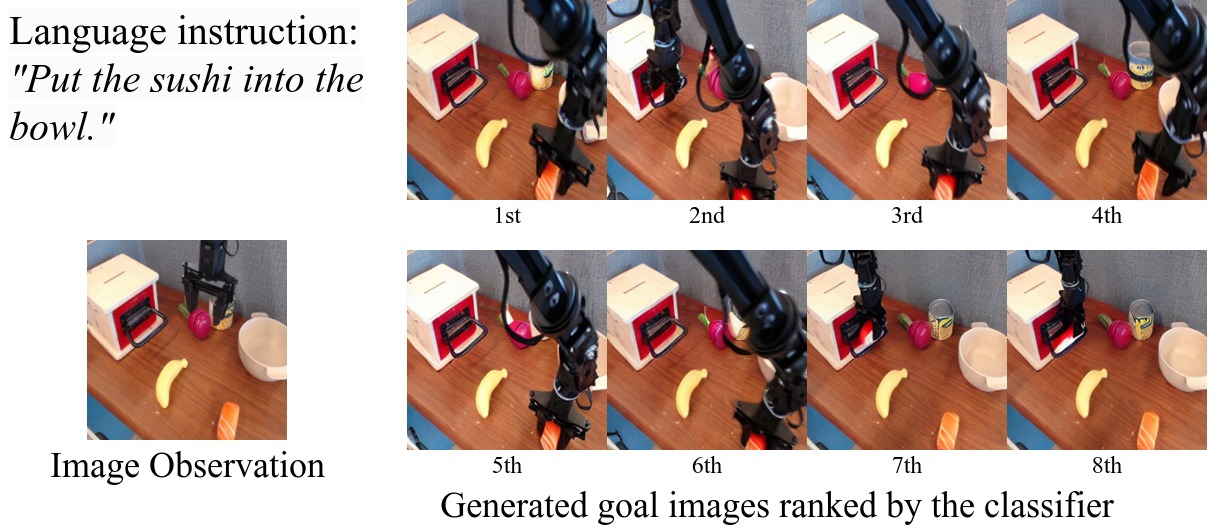}
        \vspace{1mm}
        \captionsetup{font=small}
        \caption{
            \textbf{Correct Example of Classifier Filtering}  
            The classifier correctly ranks the subgoal images where the robot is grasping the sushi higher than the subgoal images where the robot is grasping the drawer handle.
        }
        \label{fig:classifier_ranking_examples3}
    \end{subfigure}

    \centering
    \begin{subfigure}[b]{\linewidth}
        \centering
        \includegraphics[width=\linewidth]{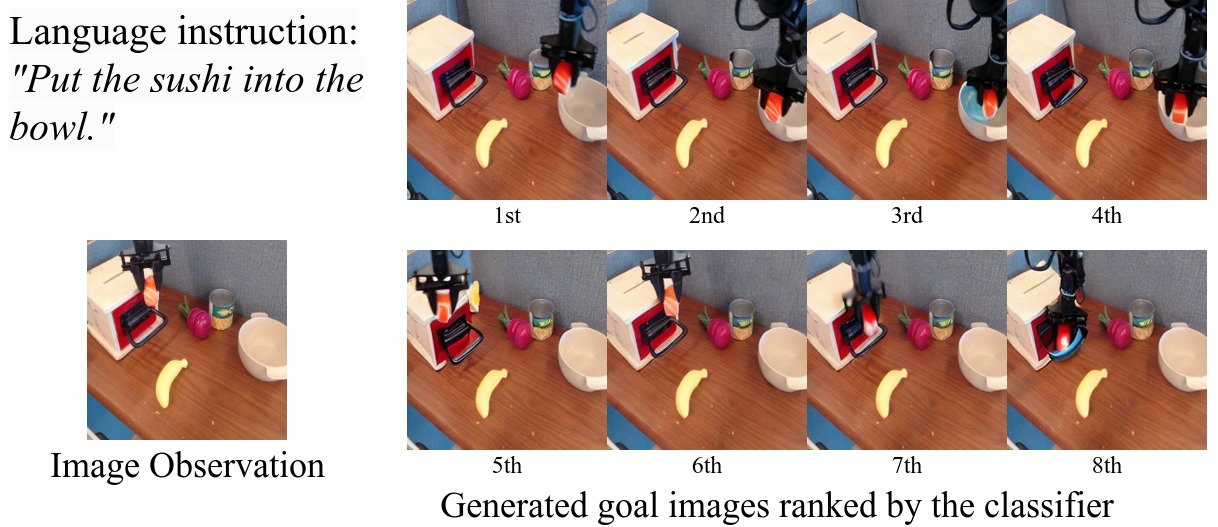}
        \captionsetup{font=small}
        \caption{
            \textbf{Correct Example of Classifier Filtering} 
            The classifier correctly ranks the subgoal images where the robot moves to place the grasped sushi into the bowl higher than the subgoal images where the robot moves its gripper towards the drawer handle. It ranks the subgoal image with the hallucinated blue bowl-like artifact last.
        }
        \label{fig:classifier_ranking_examples4}
    \end{subfigure}
    \vspace{2mm}

\end{figure}
\clearpage

\begin{figure}[h!]
    \ContinuedFloat    

    \begin{subfigure}[t]{\linewidth}
        \centering
        \includegraphics[width=\linewidth]{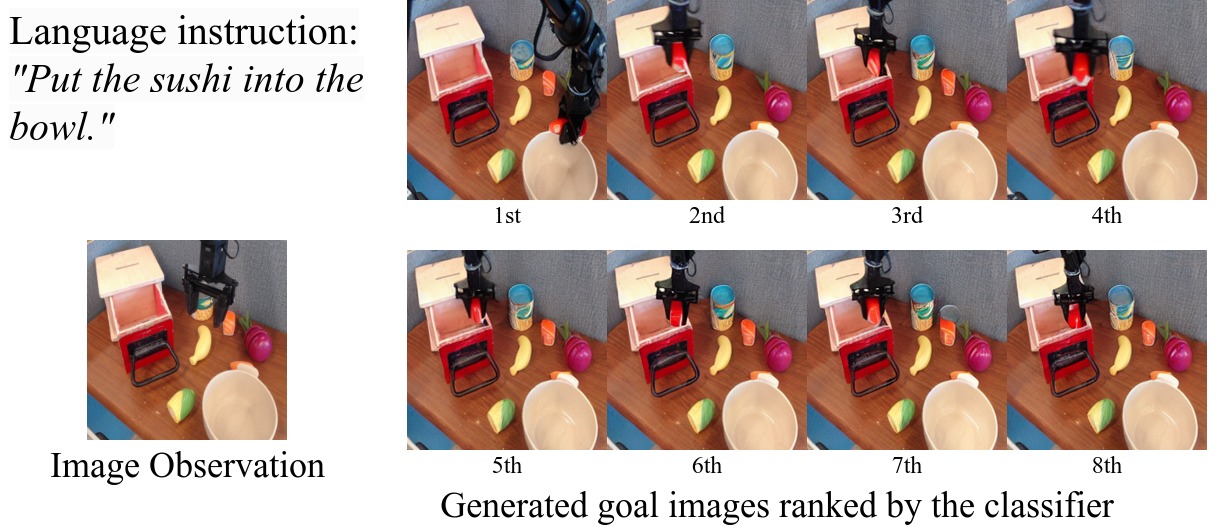}
        \captionsetup{font=small}
        \caption{
            \textbf{Correct Example of Classifier Filtering} 
            The classifier correctly ranks the subgoal image highest that shows the robot completing the correct task -- only a single generated subgoal image shows the robot placing the sushi into the bowl, while all other generated subgoal images show the robot placing the sushi into the drawer.
        }
        \label{fig:classifier_ranking_examples5}
    \end{subfigure}
    \vspace{2mm}
    
    \begin{subfigure}[b]{\linewidth}\ContinuedFloat
        \centering
        \includegraphics[width=\linewidth]{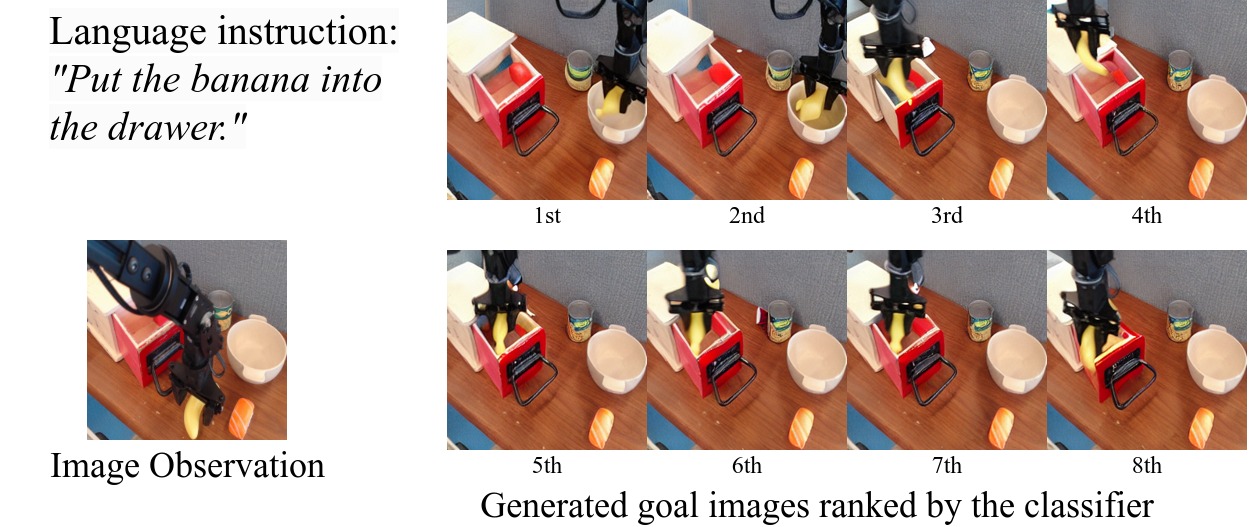}
        \captionsetup{font=small}
        \caption{
            The classifier incorrectly ranks the subgoal images higher where the robot is placing the banana into the bowl than it ranks the subgoal images where the robot is placing the banana into the drawer. This could be due to there being a strong bias for placing objects in bowls in the Bridge V2 training data.  
        }
        \label{fig:classifier_ranking_examples2}
    \end{subfigure}
\end{figure}
\clearpage

\clearpage

\subsubsection{Qualitative Analysis of Augmentation De-synchronization}

\begin{wrapfigure}[18]{R}{0.5\textwidth}
    \vspace{-1.5em}
    \centering
    \includegraphics[width=\linewidth]{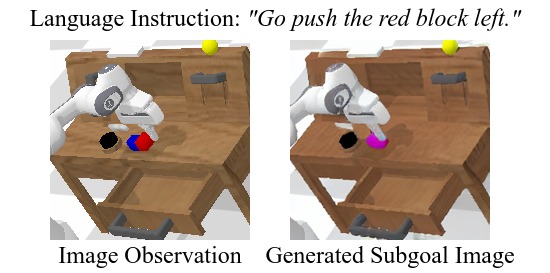}
    \captionsetup{font=small}
    \caption{
        \textbf{Generated Subgoal Image on CALVIN} A subgoal image generated by the SuSIE video model on the unseen environment D of the CALVIN benchmark. The colors and shapes of objects are different in each of the four CALVIN environments, and since the model was not trained on data from environment D, it often generates images with incorrect shapes and colors. Augmentation de-synchronization is important for the low-level policy and classifier to be able to handle these mismatches between image observations and corresponding generated subgoal images.
    }
    \label{fig:calvin_goal_image}
\end{wrapfigure}

We see that when applying augmentation de-synchronization, the number of failures due to low-level policy errors (missed grasps, dropping held objects, etc.) decreases, indicating that augmentation de-synchronization is important for the low-level policy to be able to correctly interpret and follow the subgoal images generated by the video prediction model. This is particularly important in domains where there is a large visual generalization gap between the training data and the evaluation tasks. For example, in the CALVIN benchmark, the colors and shapes of objects differ between the training and evaluation scenes. This difference causes the subgoals generated by the video prediction model to often contain objects with incorrect shapes and colors (Figure \ref{fig:calvin_goal_image}). Augmentation de-synchronization seems to be critical to allowing the low-level policy to be robust to these hallucinations and artifacts.

\subsection{Additional Ablation Experiments}
\label{appendix:ablations}

\subsubsection{Effect of Augmentation Desynchronization}

We ablate the different components of \algname when applied to SuSIE in the CALVIN benchmark (\cref{table:calvin_susie_exps}). 
Removing the augmentation desynchronization from only the low-level policy results in similar performance to base-SuSIE and \algname (SuSIE) with the augmentation desynchronization removed from both the low-level policy and the classifier. This suggests that the low-level policy performance of SuSIE without augmentation desynchronization is a significant bottleneck for SuSIE--even when selecting better subgoals via the use of filtering, performance cannot increase if the low-level policy cannot reliably reach those goals. 
Conversely, removing the augmentation desynchronization from only the classifier results in similar performance to \algname (SuSIE) without subgoal filtering. 
This suggests that, like the low-level policy, augmentation desynchronization is important for the classifier network to correctly perform its function
in \algname (SuSIE).

\begin{table}[h!]
\centering
\setlength\tabcolsep{5.2pt}
\resizebox{0.9\textwidth}{!}{%
\begin{tabular}{clcccccc}
\toprule
 & \multicolumn{1}{c}{} & \multicolumn{5}{c}{Tasks completed in a row} \\ \cmidrule(lr){3-8}
 & \multicolumn{1}{c}{\multirow{-2}{*}{Method}}  & 1  & 2  & 3  & 4  & 5  & Avg. Len.\\ \midrule
 & SuSIE & 89.8\% & 75.0\% & 57.5\% & 41.8\% & 29.8\% & 2.94 \\ %
& \algname (SuSIE)  - Aug De-sync Only & 95.2\% & 84.0\% & 69.5\% & 56.0\% & 46.2\% & 3.51 \\
& \algname (SuSIE) - Subgoal Filtering Only & 88.5\% & 75.5\% & 56.2\% & 43.0\% & 32.5\% & 2.96 \\
& \algname (SuSIE) - w/o Aug De-sync on policy & 91.5\% & 74.2\% & 56.0\% & 41.2\% & 29.8\% & 2.93 \\
& \algname (SuSIE) - w/o Aug De-sync on classifier & 95.0\% & 86.2\% & 70.0\% & 57.8\% & 47.0\% & 3.56 \\
& \algname (SuSIE) & \cellcolor[HTML]{B7E1CD}\textbf{95.2\%} & \cellcolor[HTML]{B7E1CD}\textbf{88.5\%} & \cellcolor[HTML]{B7E1CD}\textbf{73.2\%} & \cellcolor[HTML]{B7E1CD}\textbf{62.5\%} & \cellcolor[HTML]{B7E1CD}\textbf{49.8\%} & \cellcolor[HTML]{B7E1CD}\textbf{3.69} \\
\bottomrule
\end{tabular}
}
\caption{\textbf{Effect of Augmentation Desynchronization in \algname (SuSIE)}
Success rates on the validation tasks from the D environment of the CALVIN Challenge averaged across 4 random seeds. Results are shown comparing the performance of SuSIE, \algname (SuSIE), and ablations of \algname (SuSIE). \textit{\algname (SuSIE) - Aug De-sync Only} is \algname without applying subgoal filtering, \textit{\algname (SuSIE) - Subgoal Filtering Only} is \algname without applying augmentation de-synchronization to either the low-level policy or the subgoal classifier, \textit{\algname (SuSIE) - w/o Aug De-sync on policy} is \algname without applying augmentation de-synchronization on the low-level policy, and \textit{\algname (SuSIE) - w/o Aug De-sync on classifier} is \algname without applying augmentation de-synchronization on the subgoal classifier.
}
\captionsetup{font=small}

\label{table:calvin_susie_exps}
\end{table}

\subsubsection{Number of Candidate Subgoals}

We conduct an ablation over the number of candidate subgoals used for subgoal filtering in \algname (SuSIE) in the CALVIN benchmark (\cref{table:ablations_num_samples}). We find that \algname (SuSIE) achieves similar performance whether 4, 8, or 16 candidate subgoals are used. In our main results (\cref{subsec:exp_results}), we report the performance of \algname (SuSIE) on the CALVIN benchmark when using 8 candidate subgoals for filtering. For \algname (UniPi) on the CALVIN benchmark, we use 4 candidate subgoals for filtering, due to the increased computation burden of generating video subgoals with the UniPi video model vs. generating image subgoals with the SuSIE image model. In our physical experiments, we run \algname (SuSIE) using 4 candidate subgoals for filtering. 

\afterpage{
\begin{table}[htbp]
\centering
\setlength\tabcolsep{5.2pt}
\resizebox{0.8\textwidth}{!}{%
\begin{tabular}{clcccccc}
\toprule
 & \multicolumn{1}{c}{} & \multicolumn{5}{c}{Tasks completed in a row} \\ \cmidrule(lr){3-8}
 & \multicolumn{1}{c}{\multirow{-2}{*}{Method}}  & 1  & 2  & 3  & 4  & 5  & Avg. Len.\\ \midrule
& \algname (SuSIE) - 4 samples & 95.2\% & 86.0\% & 71.2\% & 60.5\% & 50.0\% & 3.63 \\
& \algname (SuSIE) - 8 samples& \cellcolor[HTML]{B7E1CD}\textbf{95.2\%} & \cellcolor[HTML]{B7E1CD}\textbf{88.5\%} & \cellcolor[HTML]{B7E1CD}\textbf{73.2\%} & \cellcolor[HTML]{B7E1CD}\textbf{62.5\%} & \cellcolor[HTML]{B7E1CD}\textbf{49.8\%} & \cellcolor[HTML]{B7E1CD}\textbf{3.69} \\
& \algname (SuSIE) - 16 samples & 95.0\% & 86.5\% & 72.8\% & 60.8\% & 48.0\% & 3.63 \\
\bottomrule
\end{tabular}
}
\captionsetup{font=small}
\caption{\textbf{Effect of Number of Candidate Goal Images Sampled in \algname (SuSIE)} Success rates on the validation tasks from environment D of the CALVIN Challenge when using \algname (SuSIE) when using 4, 8, or 16 candidate goal images with classifier filtering.
 Results are averaged across 4 random seeds. Results are similar across all numbers of samples, with 8 samples performing the best by a slight margin.}
\label{table:ablations_num_samples}
\end{table}
}

\end{document}